*Article*

# Coarse-to-Fine Structure-Aware Artistic Style Transfer

Kunxiao Liu [1], Guowu Yuan [1,*], Hao Wu [1] and Wenhua Qian [1]

1 School of Information Science and Engineering, Yunnan University, Kunming 650504, China; liukunxiao@mail.ynu.edu.cn (K.L.); gwyuan@ynu.edu.cn (G.Y.)
* Correspondence: gwyuan@ynu.edu.cn (G.Y.)

**Abstract:** Artistic style transfer aims to use a style image and a content image to synthesize a target image that retains the same artistic expression as the style image while preserving the basic content of the content image. Many recently proposed style transfer methods have a common problem; that is, they simply transfer the texture and color of the style image to the global structure of the content image. As a result, the content image has a local structure that is not similar to the local structure of the style image. In this paper, we present an effective method that can be used to transfer style patterns while fusing the local style structure into the local content structure. In our method, different levels of coarse stylized features are first reconstructed at low resolution using a Coarse Network, in which style color distribution is roughly transferred, and the content structure is combined with the style structure. Then, the reconstructed features and the content features are adopted to synthesize high-quality structure-aware stylized images with high resolution using a Fine Network with three structural selective fusion (SSF) modules. The effectiveness of our method is demonstrated through the generation of appealing high-quality stylization results and a comparison with some state-of-the-art style transfer methods.

**Keywords:** image processing; non-photorealistic rendering (NPR); style transfer; structure-aware; deep learning

## 1. Introduction

Artistic style transfer is an attractive image processing technique that is used to generate a new image that preserves the structure of a content image but carries the pattern of a style image. Recently, the seminal image-optimization method proposed by Gatys et al. [1] was used to achieve style transfer by adopting the correlation of features extracted from a pretrained deep neural network and the iterative optimization process. Like the method presented by Gatys et al. [1], style transfer by relaxed optimal transport and self-similarity (STROTSS) [2] is also an image-optimization style transfer method; this method has achieved superior stylization results by adopting the relaxed Earth mover's distance (rEMD) loss in a multiscale optimization process. However, the expensive computational cost of these image-optimization methods restricts their use in practice applications in industry. To speed up the optimization procedure, Johnson et al. [3] and Ulyanov et al. [4] proposed model-optimization style transfer methods. They train a feed-forward neural network that can be used to synthesize images with a single given style image in real time. Both adaptive instance normalization (AdaIN) [5] and whitening and coloring transforms (WCT) [6] are model-optimization methods but are also arbitrary style transfer methods, in which style patterns of arbitrary style images are transferred by adopting some feature transforms. After reviewing these methods, we have found that although local style texture and content structures can generally be combined, some key structures of the style image are not accurately learned. For example, the color blocks and brushstrokes that constitute the main objects in style images are not transferred very well. Meanwhile, in some cases, these methods produce distorted objects and incongruous artistic effects in stylized images. Therefore, our main task is to transfer the local structure of the style image to the content image and adopt a coarse-to-fine strategy to enhance the artistic details of the stylization results.

We propose a novel artistic style transfer network for fusing an essential style structure into a content structure and synthesizing a structure-aware stylized image. In our model, a Coarse Network is designed to obtain reconstructed coarse stylized features in the first stage. Because the Coarse Network only works at a low resolution, the coarse stylized features can discard trivial structure details of the content image and combine the global content structure with the style patterns. Then, the task of a Fine Network is to adopt these reconstructed coarse stylized features obtained at a low resolution and the original content image with a high resolution to synthesize the final high-resolution stylized image in the second stage. By adopting some SSF modules to fuse the coarse stylized features into the Fine Network, the final high-resolution stylized images can selectively integrate structural information from different scales. Our main contributions are as follows:

1. We introduce a novel style transfer model that can be used to synthesize appealing structure-aware stylization results. This model consists of a Coarse Network and a Fine Network. The former roughly transfers style patterns including holistic structural information and color distribution information, and then the latter enhances the details of the style patterns by fusing multiscale features.

2. We propose a SSF module for fusing the reconstructed features into the content features in Fine Network. This module can help Fine Network select essential structural information for feature fusion based on the channel attention mechanism. As a result, the color distribution of the style images can be accurately transferred.

3. It is demonstrated through experiments that our method can be used to synthesize high-quality stylizations, where the main structures of the content image are preserved and the local structures of the style image are transferred. These stylization results can maintain the same artistic expression as style images by discarding trivial content details and injecting key local style structures.

The rest of the paper is organized as follows. In Section 2, the works related to different style transfer methods are reviewed. In Section 3, the pipeline of our framework and the details of our two networks are described. Moreover, the different loss functions are introduced. Different experimental results are shown and discussed in Section 4. The conclusion is summarized in Section 5.

## 2. Related Work

### 2.1. Style Transfer

The goal of style transfer is to combine the texture of a style image with the structure of a content image. Gatys et al. [1] proposed a seminal iterative method based on a pretrained visual geometry group (VGG) network [7]. In this method, the content structure and the style texture can be used to synthesize a new image, but it is expensive and a stylized image is only generated after the training process has been completed. Inspired by Gatys et al. [1], Johnson et al. [3] proposed a feed-forward method, which can be used to synthesize arbitrary images with a fixed style by an encoder-decoder architecture; the time and computation costs are reduced when using this method. Numerous methods have been developed to speed up the style transfer process [4, 8] and improve the visual quality [9, 10, 11]. Sanakoyeu et al. [12] also improved the stylization quality by proposing a style-aware loss, but they trained a network with a set of style images instead of a style image. This approach aimed to combine many style images created by one artist to synthesize a stylized image with the overall style of this artist. Dual style generative adversarial network (DualStyle-GAN) [13] is proposed to characterize the content and style of a portrait by retaining an intrinsic style path to control the style of the original domain and an extrinsic path to model the style of the target extended domain. Peking Opera face makeup (POFMakeup) [14] also is a portrait style transfer method that can transfer the style of a portrait with a Peking Opera face to a target portrait. Lin et al. [15] combined a universal style transfer method with image fusion and color enhancement methods to solve the problems of the color scheme, the strength of style strokes and the adjustment of image contrast.

To simultaneously handle multiple styles, [16] proposed a flexible conditional instance normalization approach embedded in style transfer networks to learn multiple styles, and [17] achieved multistyle generation in a generative network architecture with a learnable inspiration layer. Ye et al. [18] adopt mechanism and instance segmentation to achieve a regional multistyle style transfer model which can solve the problem of unnatural connections between regions. Alexandru et al. [19] combined various existing style transfer frameworks to propose a novel framework that can generate intriguing artistic stylization results by performing geometric deformation and using different styles from multiple artists.

In AdaIN [5], adaptive instance normalization is implemented to train a network with various styles, providing the ability to transfer arbitrary styles after the training process. In WCT [6], the whitening and coloring transforms are adopted to synthesize arbitrary styles with a pretrained VGG network and a series of pretrained image restructuring decoders. Based on WCT, Wang et al. [20] achieved the diversity of style transfer by adopting a deep feature perturbation (DFP) operation while preserving the quality of stylization results, and Wang et al. [21] synthesized ultra-resolution stylized images and reduced the convolutional filters by a knowledge distillation method. Style-attentional network (SANet) [22] is also an arbitrary style transfer method that can be used to efficiently generate stylized images by injecting local style patterns into content features based on the style attention mechanism.

### 2.2. Style Transfer Based on Multiscale Learning

Recently some style transfer methods have been used to transfer style patterns based on multiscale learning. Multiscale holistic style transfer is achieved in Avatar-Net [23] based on the use of an hourglass with multiple skip connections and a style decorator. STROTSS [2] is an image-optimization method that adopts multiscale learning to update the content image and generate high-quality stylized images. Yang et al. [24] proposed a novel video style transfer framework that can render high-quality artistic portraits based on the multiscale content features and preserve the frame details. A Laplacian pyramid style network (LapStyle) [25] also exhibits high visual quality and is based on a Drafting Network and a Revision Network. First, the former transfers the global style patterns, and then, the latter enhances local style details. However, too many content structure details are preserved in these methods. Key local style structures are not fused into stylized images in any of these methods. In contrast, our method transfers global style patterns at low resolution using a Coarse Network, which only needs to be trained once to reconstruct coarse stylized features. Our Fine Network enhances local style details with multiscale features from the Coarse Network and the high-resolution content image. As a result, our method can discard trivial local content structures and synthesize high-quality structure-aware stylized images by a coarse-to-fine process. The differences between our method and the previous studies are shown in Table 1.

**Table 1.** The differences between our method and the previous studies.

| Methods | Image-optimization | Model-optimization | Single style | Multiple style | Arbitrary style |
|---|---|---|---|---|---|
| Ours | | √ | √ | | |
| [1, 2] | √ | | √ | | |
| [3, 4, 8, 9, 10, 11, 12, 13, 14, 15, 24, 25] | | √ | √ | | |
| [16, 17, 18, 19] | | √ | | √ | |
| [5, 6, 20, 21, 22, 23] | | √ | | | √ |

## 3. Proposed Method

### 3.1. Framework Overview

Inspired by the painting process of artists, in which the coarse structure and color distribution are first constructed and then fine details are added, our framework employs a Coarse Network and Fine Network to simulate the coarse-to-fine process. As shown in Figure 1, given a content image $x_c \in \mathbb{R}^{3 \times h_c \times w_c}$ and a style image $x_s \in \mathbb{R}^{3 \times h_s \times w_s}$, our model eventually generates a stylized image $x_{cs} \in \mathbb{R}^{3 \times h_{cs} \times w_{cs}}$. In the first stage, the Coarse Network takes $\bar{x}_c$ and $\bar{x}_s$ as inputs, where $\bar{x}_c$ and $\bar{x}_s$ are the results of downsampling $x_c$ and $x_s$ by 2, respectively. Then three different restructured coarse stylized features $\bar{f}_r^{(i)} \in \mathbb{R}^{c_r^{(i)} \times h_r^{(i)} \times w_r^{(i)}}$ ($i$ = 1, 2, 3) are generated by the Coarse Network, where $c_r^{(i)}$, $h_r^{(i)}$, and $w_r^{(i)}$ are the number of channels, height, and width of the $i$ th restructured feature, respectively. In the second stage, the Fine Network takes $x_c$ and $\bar{f}_r^{(i)}$ as inputs and then generates the final stylized image $x_{cs}$ by adopting SSF modules for feature fusion.

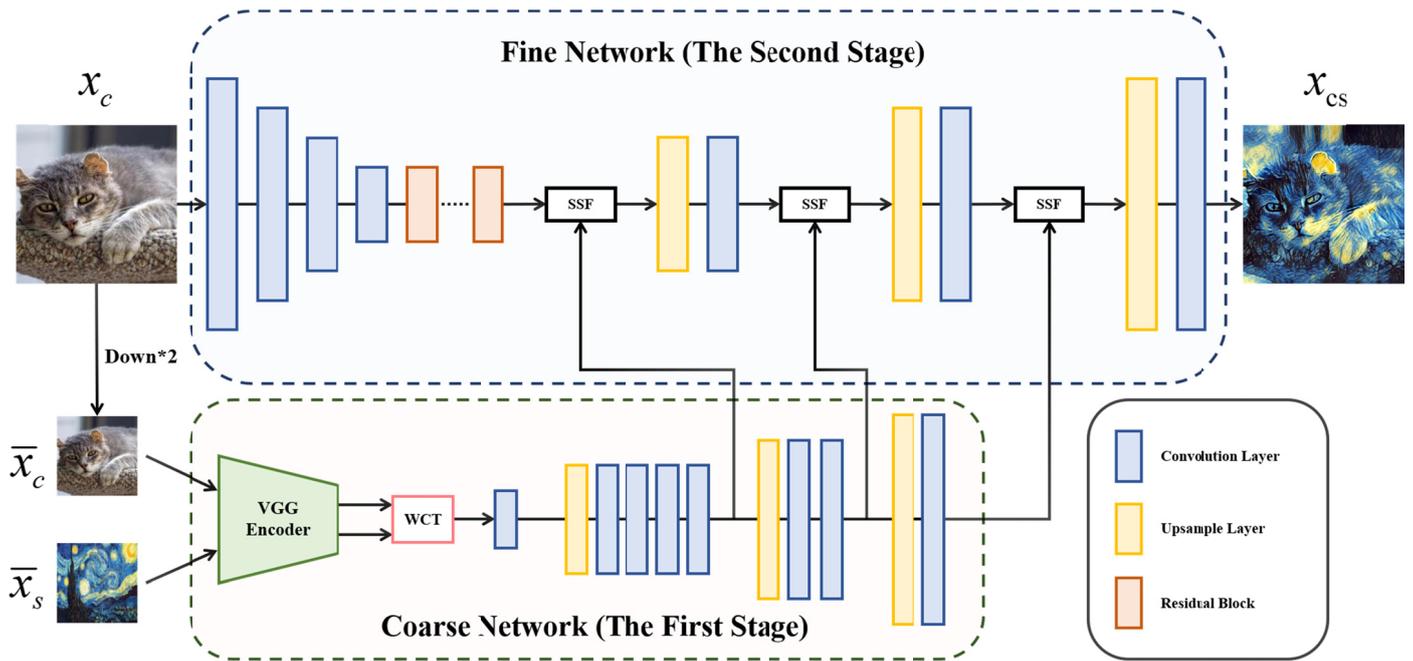

**Figure 1.** Overview of our framework.

As shown in Figure 2, the different stylized images are generated by our method. In Figure 2(**b**), we only adopt the last restructured coarse stylized features to directly restructure the coarse stylized image by the Coarse Network in the first stage. The coarse stylized image discards unnecessary local structures of the content image and transfers the global color distribution of the style image. Then, the Fine Network is employed to encode the high-resolution content image to obtain the content features, and these content features and three different coarse reconstructed features from the Coarse Network are decoded to generate the high-quality structure-aware stylized image in the second stage. As shown in Figure 2(**c**), the final appealing stylized image is synthesized by adopting our full model with a Coarse Network and a Fine Network. Moreover, to show the local style structure of the final stylized image more clearly, we use the color control method [26] to keep the color of the final stylized image consistent with the color of the original content image. As illustrated in Figure 2(**d**), although the color distribution of the stylized image remains the same as that of the content image, the local structure of the stylized image is similar to that of the style image.

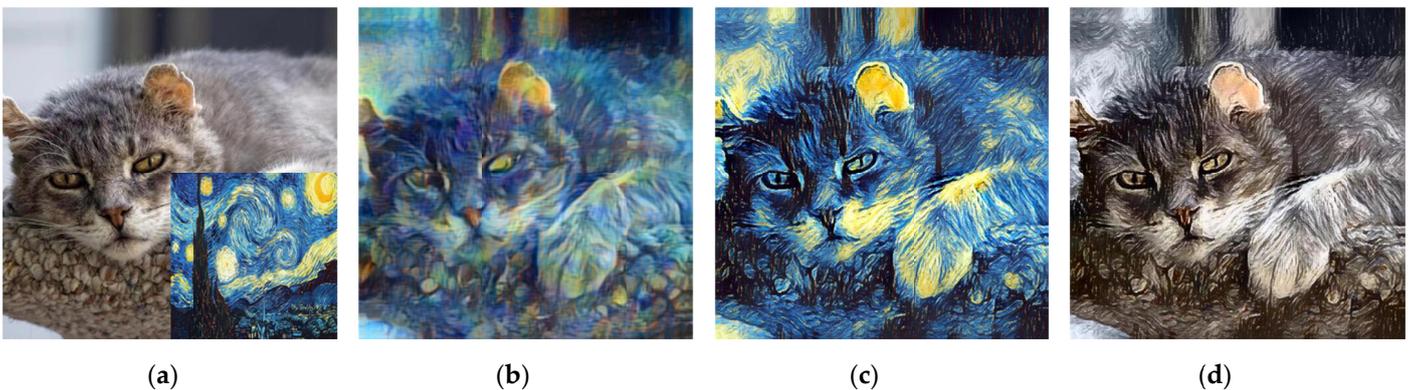

**Figure 2.** Different stylization results from the same content image and style image: (**a**) The content image is a cat, and the style image is Starry Night by Vincent van Gogh; (**b**) This stylized image is directly generated by our Coarse Network in the first stage; (**c**) The final stylized image is generated by our full model in the second stage; (**d**) This stylized image maintains the same color as the content image using color control.

*3.2 Coarse Network*

One problem with recent style transfer methods is that too many structural details of the content image are retained during the transfer of style patterns. In the stylized image, there are some small structures from the content image that do not change; they simply transfer the color and texture of the style image. These local structures that do not exist in the style image appear in the stylized image, resulting in a stylized image that fails to show the spirit of the artistic expression of the style image. The reason is that these methods directly extract features from high-resolution images and cannot decide which details to discard from the content image. Contrary to previous work, our Coarse Network transfers rough style patterns at low resolution. As a result, there is a larger receptive field to learn low-frequency information to determine the overall structure of the image. Then, some unnecessary high-frequency information is ignored during training. As shown in Figure 3, the Coarse Network can transfer more details that are unnecessary in the coarse stylized image at high resolution. At low resolution, the Coarse Network can discard some trivial details of the structure and keep the objects smooth in the stylized image.

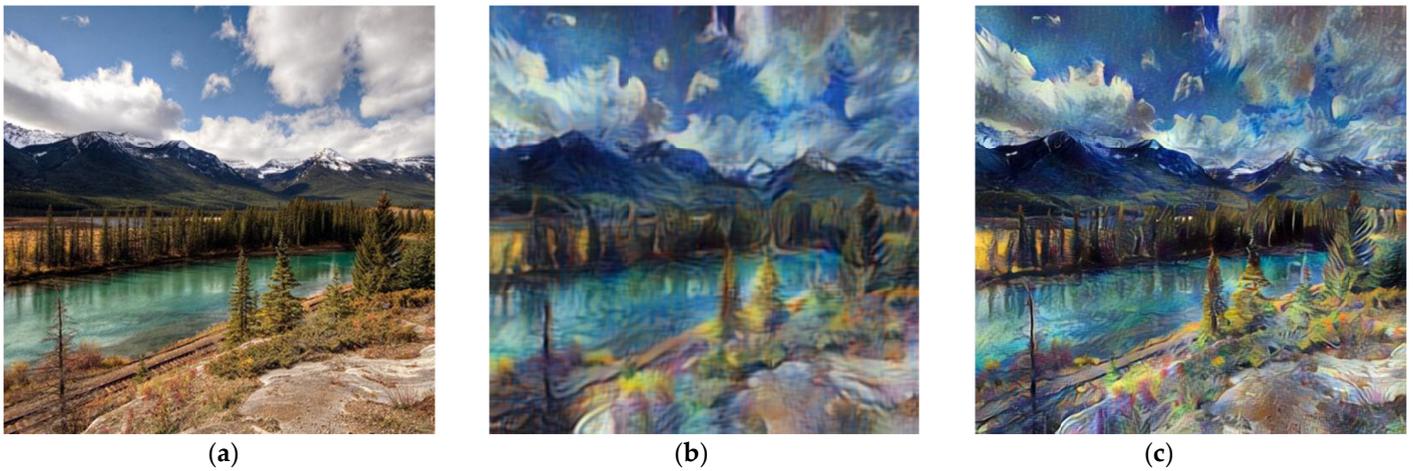

**Figure 3.** Comparison of two stylized images generated by the Coarse Network at different resolutions: (**a**) The original content image with resolution of 512 × 512; (**b**) The stylized image with resolution of 256 × 256; (**c**) The stylized image with resolution of 512 × 512.

3.2.1 WCT Module

Inspired by WCT [6], our Coarse Network adopts whitening and coloring transforms to transfer coarse style patterns at low resolution. The whitening transform can remove inessential information related to style while preserving the global structure of the content. Then, the coloring transform can capture the salient visual style and fuse some style structures into content structures. WCT is a multilevel stylization process that uses different rectified linear unit (ReLU) layers of VGG features ReLU_X_1 (X = 1, 2, …, 5) and transfers style patterns in a coarse-to-fine pipeline. The higher layer features are adopted to capture complex local structures, while lower layer features carry low-level color and texture information. The difference between our Coarse Network and WCT is that we only use a single-level whitening and coloring transform for stylization. Moreover, we do not directly reconstruct the stylized features to generate an image; however, we utilize the reconstructed features at different layers during reconstruction. As a result, our Coarse Network, which has the ability to capture the multilevel information by reconstructing the coarse stylized features at different levels, can save computing resources.

3.2.2 Architecture of Coarse Network

The architecture of Coarse Network, which is shown in Figure 1, includes an encoder, a WCT module, and a decoder. (1) The encoder is a pretrained VGG-19 network, which is fixed during training. Given $\bar{x}_c$ and $\bar{x}_s$, the VGG encoder extracts the content feature $\bar{f}_c$ and the style feature $\bar{f}_s$ at ReLU_4_1. (2) Then, we apply a WCT module for whitening and coloring transformation. As shown in Figure 4(**a**), the whitening transform is adopted to linearly transform $\bar{f}_c$ to obtain $\bar{f}_c'$. Next, the coloring transform is carried out to obtain $\bar{f}_{cs}$ by using $\bar{f}_c'$ and $\bar{f}_s$. (3) Finally, we adopt a reconstruction decoder to reconstruct the coarse stylized feature $\bar{f}_{cs}$. The decoder is designed to be symmetrical to the VGG-19 network, with the nearest neighbor upsampling layer used for enlarging the feature map. We take $\bar{f}_{cs}$ as input for reconstruction and then generate these restructured stylized features $\bar{f}_r^{(i)}$ as outputs. In this reconstruction decoder, these outputs are output before the second upsampling layer, the third upsampling layer, and after the last convolution layer. These $\bar{f}_r^{(i)}$ will become a part of the input of the Fine Network.

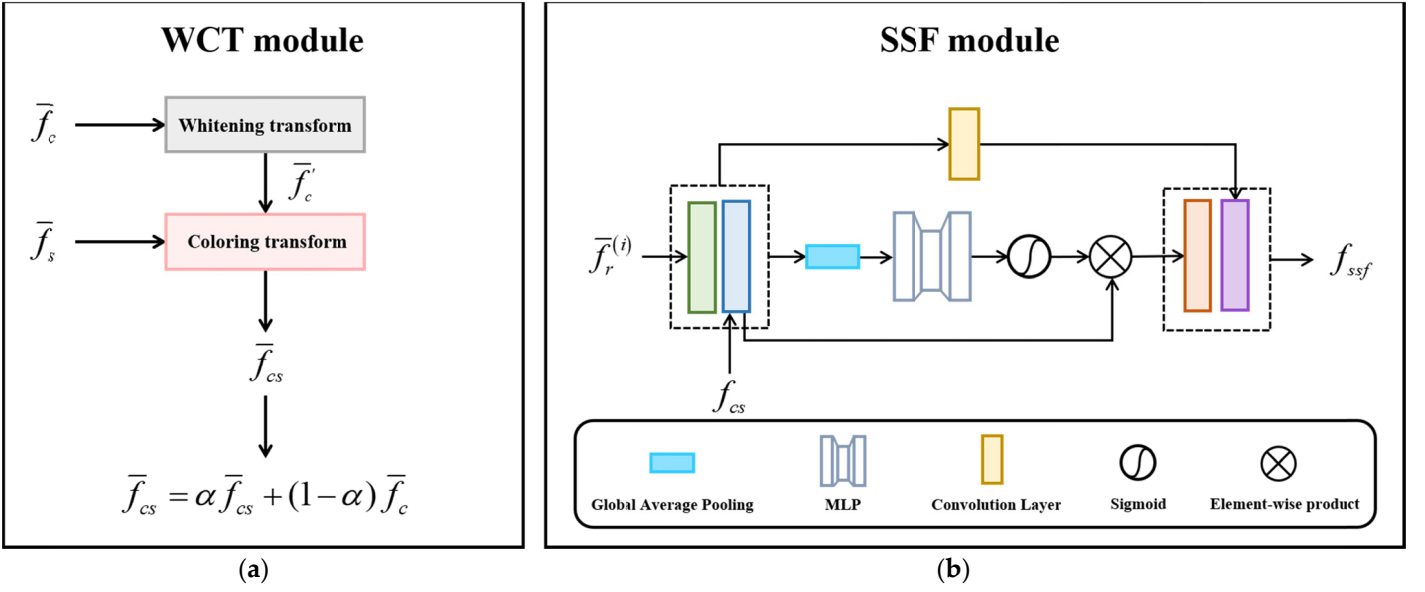

**Figure 4.** The schematics for two modules: (**a**) WCT module; (**b**) SSF module.

*3.3 Fine Network*

The Fine Network aims to synthesize high-resolution stylized images by fusing the reconstructed coarse stylized features into the reconstructed content features. The reconstructed content features are from the high-resolution content image, and contains the global semantic information and local detail information. Contrary to the reconstructed content features at high resolution, the reconstructed stylized features generated from the Coarse Network preserve only the main content structure while blending some local structural style information. By fusing multiscale information, the Fine Network can pay more attention to the holistic structure of the content and ignore some trivial details based on our SSF modules. Then, some significant details can be added to the structure and appealing artistic effects in the stylized image can be enhanced. In addition, fusing the reconstructed coarse stylized information can greatly reduce the time cost of the training process of the Fine Network, and the desired stylization results can be achieved at an earlier point in time.

3.3.1 SSF Module

The structural selective fusion (SSF) module is designed to fuse the reconstructed coarse stylized features from the Coarse Network into the reconstructed content features in the decoder of the Fine Network. Inspired by the attention mechanism [27, 28], we employ a weight matrix to select key structural information of the reconstructed content features, which is learned by adopting the merged features. The merged features are obtained by concatenating reconstructed coarse stylized features and the reconstructed content features. The matrix can help the SSF module obtain the selective features that focus on meaningful structural information, and the selective feature is one part of the output of the SSF module. Another part of the output is the refined merged features that include different scale information such as some crucial local textures or global structures.

The architecture of the SSF module is shown in Figure 4(**b**). First, we concatenate the reconstructed coarse stylized features $\bar{f}_r^{(i)}$ and the reconstructed content features $f_{cs}$ as input $f_{csr} \in \mathbb{R}^{(c_{cs}+c_r) \times w_r \times h_r}$. The reconstructed content features $f_{cs}$ are the output of the convolution layer in the decoder of the Fine Network (except that the first SSF module uses the content features $f_c$ from the encoder of the Fine Network as $f_{cs}$). We adopt an average-pooling operation to aggregate the spatial information of $f_{csr}$ to generate the input of the multilayer perceptron, which is adopted to produce an attention map $M_{cs} \in \mathbb{R}^{c_{cs} \times 1 \times 1}$ as the weight matrix. In summary, the attention map is calculated as:

$$M_{cs}(f_{csr}) = \sigma\left(MLP\left(AvgPool(f_{csr})\right)\right) \tag{1}$$

where $\sigma$ denotes the sigmoid function. Then the selective feature $f'_{cs}$ is calculated as:

$$f'_{cs} = M_{cs}(f_{csr}) \otimes f_{cs} \tag{2}$$

where $\otimes$ denotes elementwise multiplication. Meanwhile, $f_{csr}$ is fed into a convolutional layer to produce a refined merged feature $f'_{csr} \in \mathbb{R}^{c_r \times w_r \times h_r}$. Eventually, the SSF module generates the final output $f_{ssf} \in \mathbb{R}^{(c_{cs}+c_r) \times w_r \times h_r}$ as the fused feature by directly concatenating $f'_{cs}$ and $f'_{csr}$.

3.3.2 Architecture of Fine Network

As shown in Figure 1, Fine Network is designed as a flexible encoder-decoder architecture, with an encoder, a series of residual blocks, and a decoder. The encoder contains a convolutional layer with a stride of 1 and three convolutional layers with strides of 2, followed by several residual blocks. The decoder contains three upsampling layers, three convolutional layers with strides of 1, and three SSF modules. We use an SSF module before each upsampling layer. Given the content image $x_c$ as the input of Fine Network, the encoder and several residual blocks generate the content feature $f_c$. Then, SSF modules generate the fused features $f_{ssf}$ by taking $\overline{f}_r^{(i)}$ and $f_{cs}$ as inputs, where $f_{cs}$ is the output of these convolution layers in the decoder (except the first SSF module, which takes $f_c$ as $f_{cs}$). These fused features $f_{ssf}$ are fed into an upsampling layer and a convolution layer. Finally, the decoder generates the final stylized image $x_{cs}$ after the last convolution layer.

3.4 Loss Function

Our Coarse Network only needs to train once, and it is fixed during the training of the Fine Network. Compared to WCT [6], we only train one reconstruction decoder network to reconstruct the coarse stylized feature. Our Coarse Network can reconstruct the stylized features at three different levels or directly generate a coarse stylized image by taking advantage of the reconstruction decoder. Following WCT, we adopt pixel reconstruction and perceptual loss [3] to train our decoder for image reconstruction,

$$l_{re} = \| I_o - I_i \|_2^2 + \lambda \| \Phi(I_o) - \Phi(I_i) \|_2^2 \qquad (3)$$

where $I_i$ and $I_o$ are the input image and output image, respectively, and $\Phi$ is the VGG encoder that extracts features at ReLU_X_1 (X = 1, 2, 3, 4). In addition, $\lambda$ is the weight to balance the two losses.

The Fine Network is optimized with content and style loss during training. As shown in Figure 5, we keep a single $x_s$ and a set of $x_c$ from a content dataset, then $x_{cs}$ is a stylized image generated by the Fine Network. For $x_s$, $x_c$ and $x_{cs}$, we can use a pretrained VGG-19 encoder to extract their features $F_c^{(t)} \in \mathbb{R}^{c_t \times h_t \times w_t}$, $F_s^{(t)} \in \mathbb{R}^{c_t \times h_t \times w_t}$ and $F_{cs}^{(t)} \in \mathbb{R}^{c_t \times h_t \times w_t}$, where $t$ denotes the features extracted at ReLU_$t$ ($t$ = 1_1, 1_2, 2_1, 2_2, 3_1, 3_3, 4_1).

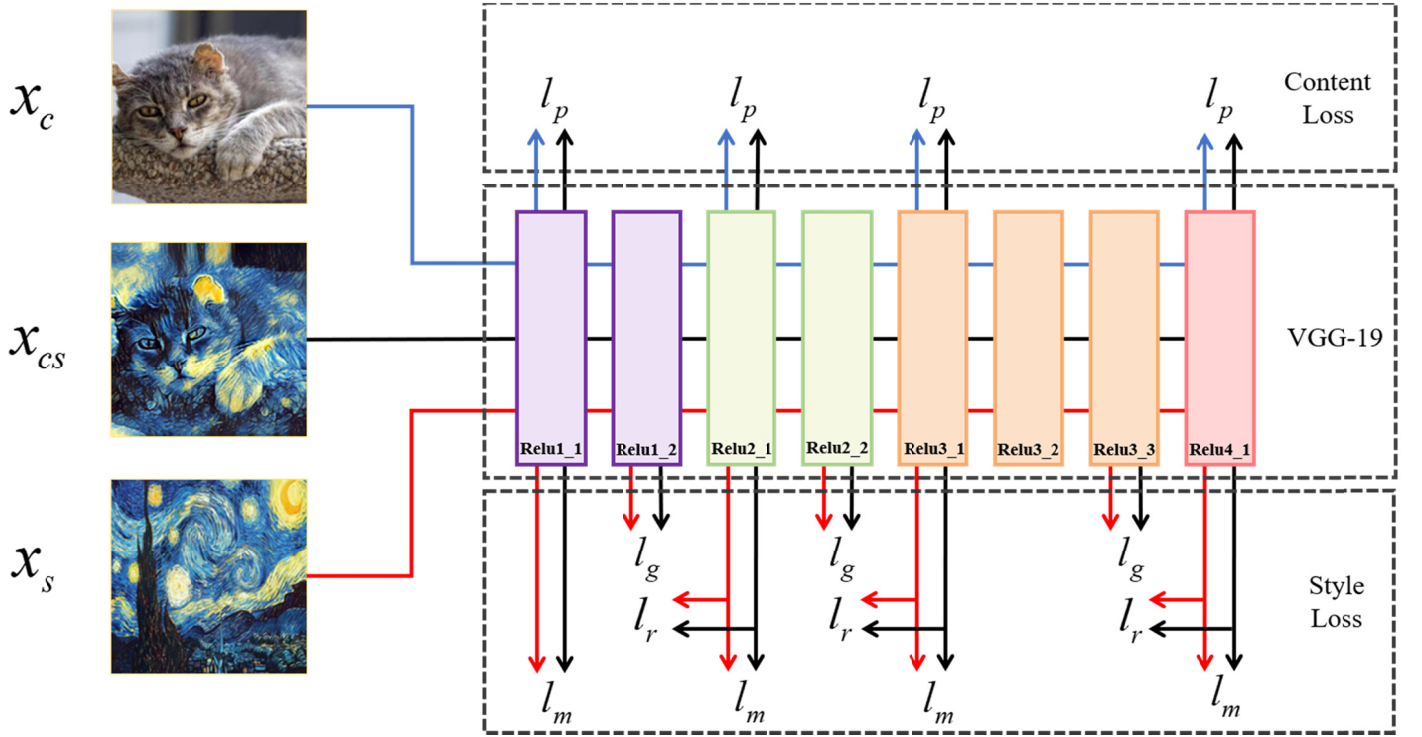

**Figure 5.** The schematic for loss network.

For content loss, we adopt the commonly used perceptual loss between $F_c^{(t)}$ and $F_{cs}^{(t)}$ proposed in [3]. The perceptual loss can measure high-level perceptual and semantic differences between images and is defined as:

$$l_p = \frac{1}{c_t h_t w_t} \| F_c^{(t)} - F_{cs}^{(t)} \|_2^2 \tag{4}$$

For style loss, we adopt three different style losses. The first and most significant style loss is the relaxed Earth mover's distance (rEMD) loss [2], which helps the Fine Network generate visual effects with minimum distortion to the layout of the content image. This loss plays a key role in migrating the structural forms of the style image to the target image. The rEMD loss between $F_s^{(t)}$ and $F_{cs}^{(t)}$ can be calculated as:

$$l_r = \max\left(\frac{1}{h_t w_t} \sum_{i=1}^{h_t w_t} \min_j C_{ij}, \frac{1}{h_t w_t} \sum_{j=1}^{h_t w_t} \min_i C_{ij}\right) \tag{5}$$

where $C$ is the cost matrix, which can be calculated as the cosine distance between $F_s^{(t)}$ and $F_{cs}^{(t)}$:

$$C_{ij} = D_{cos}\left(F_{s,i}^{(t)}, F_{cs,j}^{(t)}\right) = 1 - \frac{F_{s,i}^{(t)} \cdot F_{cs,j}^{(t)}}{\| F_{s,i}^{(t)} \| \| F_{cs,j}^{(t)} \|} \tag{6}$$

The second style loss is the commonly used style reconstruction loss proposed by Gatys et al [1], which is the difference between the Gram matrices of $F_s^{(t)}$ and $F_{cs}^{(t)}$:

$$l_g = \| G\left(F_s^{(t)}\right), G\left(F_{cs}^{(t)}\right) \|_2^2 \tag{7}$$

where $G$ denotes the calculation of the Gram matrix of the feature vectors. Finally, we use the mean-variance loss as the third style loss, which is similar to the style reconstruction loss. We can use this type of loss to reduce unnecessary visual effects in the stylized image and keep the magnitude of the stylized feature the same as that of the style feature:

$$l_m = \| \mu\left(F_s^{(t)}\right) - \mu\left(F_{cs}^{(t)}\right) \|_2^2 + \| \sigma\left(F_s^{(t)}\right) - \sigma\left(F_{cs}^{(t)}\right) \|_2^2 \tag{8}$$

where $\mu$ and $\sigma$ denote the mean and covariance of the feature vectors, respectively.

The overall optimization objective is defined as:

$$L = \alpha l_p + \lambda_1 l_r + \lambda_2 l_g + \lambda_3 l_m \tag{9}$$

where $\alpha$, $\lambda_1$, $\lambda_2$ and $\lambda_3$ are weight terms. By adjusting $\alpha$, we can control the degree of stylization. Specifically, $l_p$ and $l_m$ both work on ReLU_1_1, ReLU_2_1, ReLU_3_1 and ReLU_4_1; then, $l_r$ works on ReLU_2_1, ReLU_3_1 and ReLU_4_1. Following Johnson et al. [3], $l_g$ works on ReLU_1_2, ReLU_2_2 and ReLU_3_3.

## 4. Experimental Results and Analysis

*4.1. Experimental Dataset and Implementation Details*

During training, we use the MS-COCO [29] dataset as the set of content images and select some famous art paintings as style images. To show the experimental results of our method, we also select some copyright-free images as content images from Pexels.com.

In our experiment, the Coarse Network is trained on the MS-COCO dataset only once for image reconstruction, and the weight $\lambda$ in Equation (1) is set as 1. In the experiments, we use the content images and the style image with a resolution of 512 × 512. Then these images are downsampled by 2. The images input into the Coarse Network has a resolution of 256 × 256. During the training of the Fine Network, we use the Adam [30] optimizer with a learning rate of 1e-4, and the batch size is set as 1 because of the limitation of the graphics processing unit (GPU) memory. To train a style, a training process consists of 15,000 iterations. The loss weight terms, $\alpha$, $\lambda_1$, $\lambda_2$ and $\lambda_3$ are set to 1, 20, 1,000 and 5, respectively. The experimental environment configuration is shown in Table 2.

**Table 2.** Experimental environment configuration.

| Designation | Information |
|---|---|
| Operating system | Windows 10 |
| System configuration | CPU: AMD Ryzen 9 5900X |
| | GPU: NVIDIA GeForce RTX 3090 |
| Software | PyCharm 2021.3.1 (Community Edition) |
| | Python 3.8.12 |
| Python library | Cuda 11.7 |
| | Pytorch 1.8 |
| | Torchvision 0.9 |
| | Numpy 1.21 |
| | Matplotlib 3.5.1 |

*4.2. Qualitative Comparisons with Prior Works*

Inspired by the recent WCT [6] and STROTSS [2] methods, our method adopts the whiting and coloring transformation proposed in WCT and the rEMD loss proposed in STROTSS. In Figure 6, we compare our method with WCT and STROTSS. WCT can transfer the color distribution and simple texture of arbitrary style images; however, some context local structure is discarded, resulting in messy and disordered stylized images (e.g., rows 1, 2 and 3). STROTSS is an image-optimization style transfer method that transfers the visual attributes from the style image to the content image with minimum semantic distortion. Nevertheless, too many structural details are preserved, and the overall palette of the style image is not accurately transferred (e.g., rows 2, 3). In contrast to these two methods, our method can transfer the main structure and discard some trivial details of the content image. Moreover, some notable local structures of the style image, such as brushstrokes, can be fused into the global structure of the content image, and the overall palette of the stylized image remains the same as that of the style image. For example, in the second and fourth rows, the color blocks of mountains and the brushstrokes of vegetation in our stylized images are explicitly similar to those in style images. Our model can learn some key style structures while ignoring some unimportant content details.

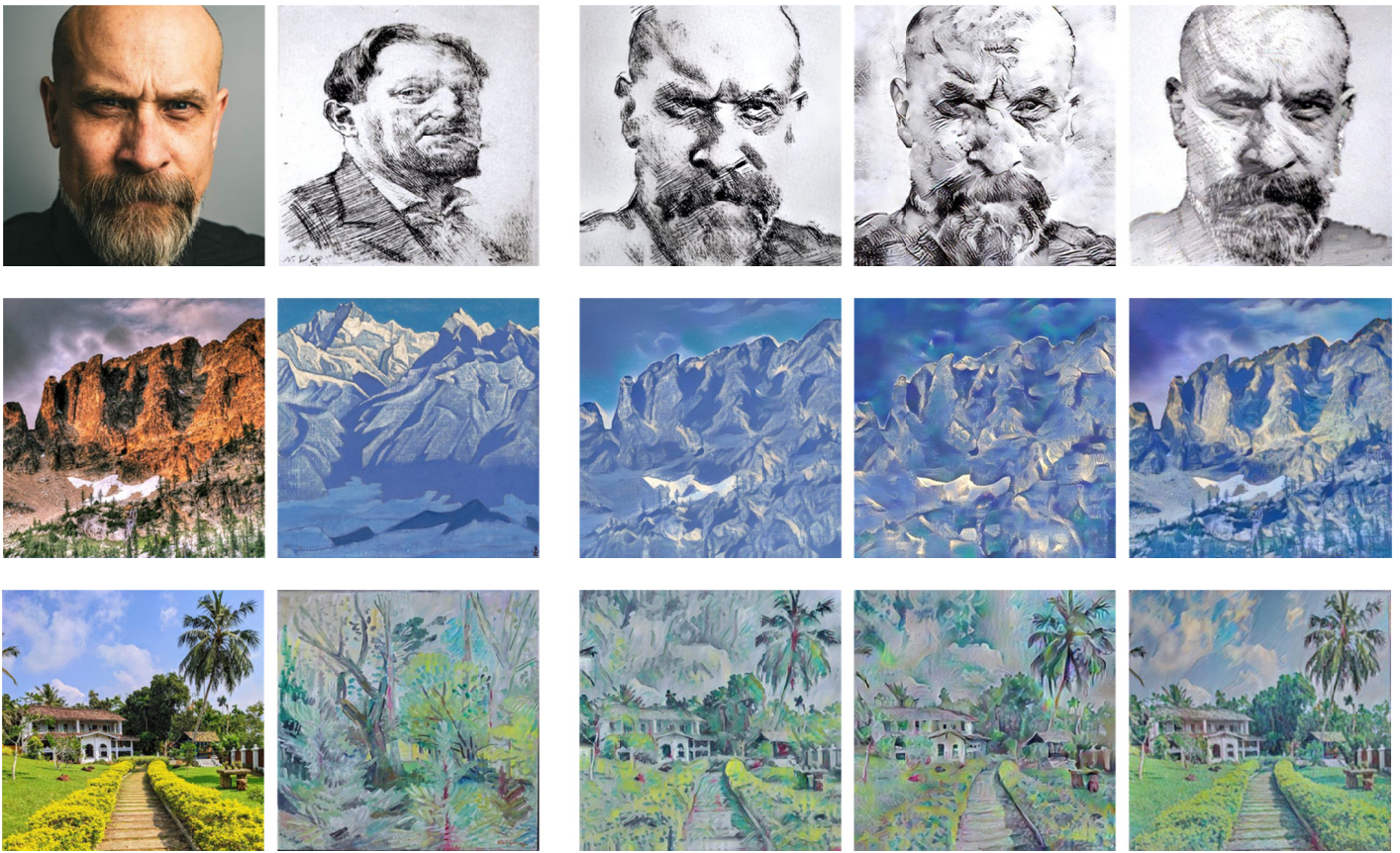

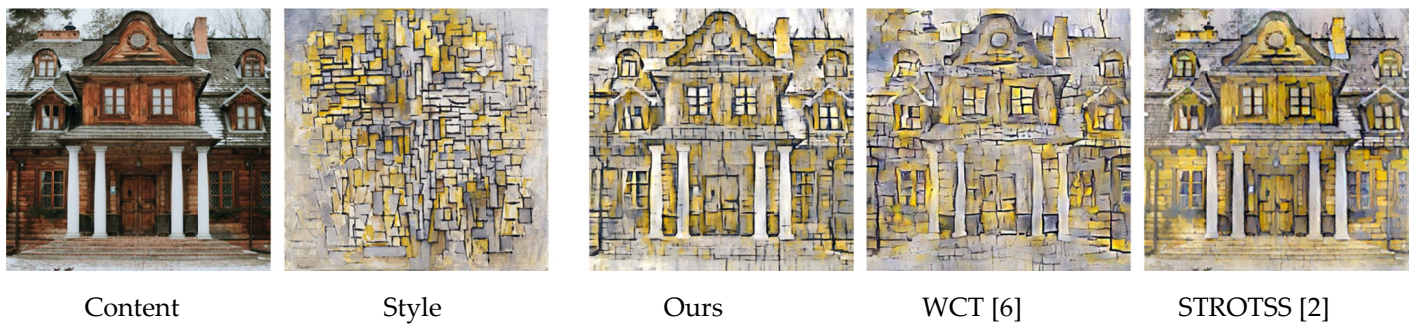

| Content | Style | Ours | WCT [6] | STROTSS [2] |

**Figure 6.** Qualitative comparisons between our method, WCT and STROTSS.

As shown in Figure 7, we compare our method with other state-of-the-art style transfer methods. Gatys et al. [1] proposed the original optimization-based style transfer algorithm, which can transfer the overall style texture and the color distribution. However, some incongruous textures appear in the stylized images leading to the stylizations looking unnatural (e.g., rows 4, 5 and 6). Similar to our method, the method proposed by Johnson et al. [3] is also a feed-forward method. It can combine the local color and texture of style images with the structure of content but often maintains too many content structures and may only play a role in shifting the color histogram in some cases (e.g., rows 1, 2 and 3). AdaIN [5] and SANet [22] are both arbitrary style transfer models, which mainly transfer simple style patterns. AdaIN often fails to transfer the color distribution of style images, and SANet has the severe problem of messy texture and disordered structure (e.g., rows 4, 5 and 6). All of the methods mentioned above maintain some unnecessary small local structures of the content images, and the essential local structures of style images are not integrated into the target image. In contrast to these methods, our model can simultaneously transfer the style color distribution accurately and combine the local style structure with the global content structure. For example, in the fourth row, the image of the rabbits generated by our method looks more harmonious and natural in the stylized image. It seems as though the style image consists of ink dots; the same artistic expression can be exhibited by our method.

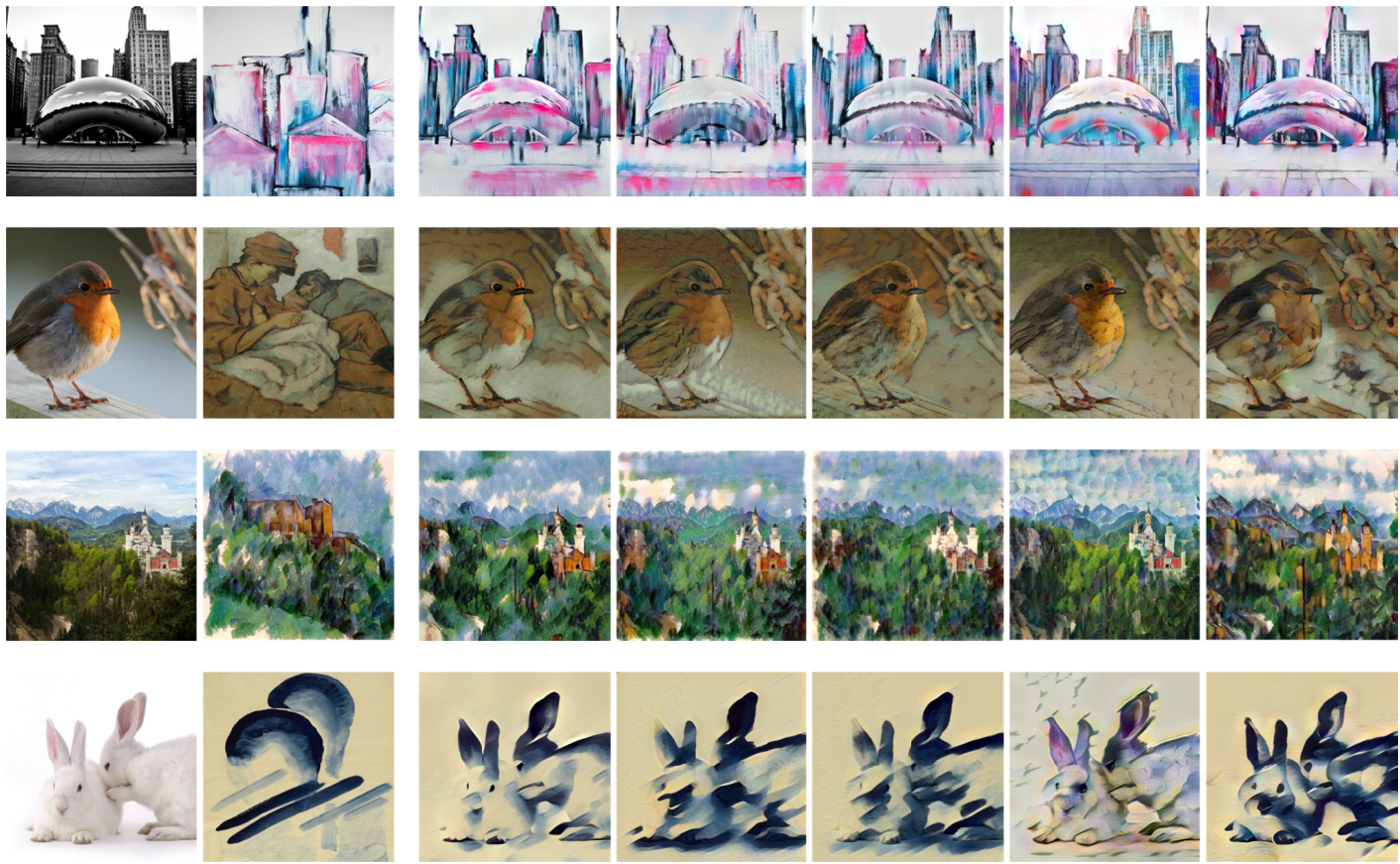

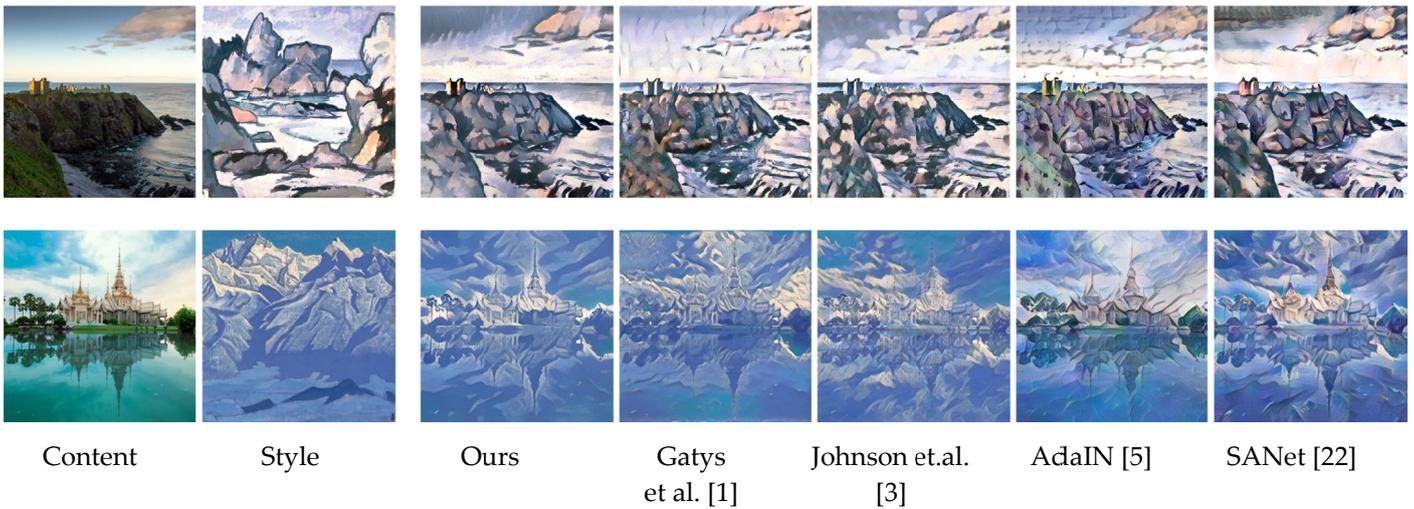

**Figure 7.** Qualitative comparisons between our method and other state-of-the-art methods.

### 4.3. Quantitative Comparisons with Prior Works

In the experiment of quantitative comparisons, we use the learned perceptual image patch similarity (LPIPS) proposed in [31] and the structural similarity index measurement (SSIM) proposed in [32] to compute the difference in style structure between the stylized image and style image. In each method, 1,500 pairs of stylized and style images that include ten styles are used to compute the averaged distance. As shown in Table 3, lower values indicate the higher similarity of human perceptual judgments when we use LPIPS as the metric, and higher values indicate the higher structural similarity when we use SSIM as the metric. For both evaluation metrics, our proposed method achieves the highest similarity in style structure. The experimental results show that our method can synthesize structure-aware stylized images that have a higher structural similarity to the style images.

**Table 3.** Quantitative comparisons of LPIPS and SSIM between our method and six state-of-the-art methods.

| Method | Our | WCT [6] | STROTSS [2] | Gatys et al. [1] | Johnson et al. [3] | AdaIN [5] | SANet [22] |
| --- | --- | --- | --- | --- | --- | --- | --- |
| LPIPS | 0.6287 | 0.6393 | 0.6516 | 0.6477 | 0.6452 | 0.6445 | 0.6408 |
| SSIM | 0.2135 | 0.1975 | 0.2108 | 0.2022 | 0.2068 | 0.1933 | 0.1893 |

### 4.4. Comparisons of Time Efficiency with Prior Works

We further compare the time efficiency of our proposed method with other state-of-art methods. In each method, we synthesize 100 stylized images with a resolution of 512 × 512. All experiments are conducted on the same environment configuration. As shown in Table 4, Johnson et al. [3] achieve the highest time efficiency because they only use a simple encoder-decoder architecture to generate stylized images. Like [3], AdaIN [5] and SANet [22] also use the simple encoder-decoder network to generate stylized images. But they apply some feature transform modules in their networks to integrate content features and style features. As a result, their time efficiencies are lower than [3] but are still satisfactory. Different from these three methods that work at the same image scale, our model includes two different networks and works in two stages. Although our model can capture richer multiscale information and synthesize higher-quality stylized images, the time efficiency of our method is only slightly lower than that of AdaIN and SANet. We traded a small increase in time cost for a promising improvement in the quality of stylized images. WCT [6] has a low time efficiency because it uses five encoders and decoders to generate a stylized image. The time efficiencies of STROTSS [2] and Gatys et al. [1] are far lower than other methods because they are image-optimization methods that only generate one stylized image after a training process.

**Table 4.** Running time comparison between our method and six state-of-the-art methods (in seconds).

| Method | Our | WCT | STROTSS | Gatys et | Johnson | AdaIN | SANet |

|  |  | [6] | [2] | al. [1] | et al. [3] | [5] | [22] |
|---|---|---|---|---|---|---|---|
| Time |  | 0.829 | 2.816 | 40.157 | 20.418 | 0.075 | 0.105 | 0.291 |

*4.5. User Study*

The user study is conducted on social media, and all participants are anonymous and voluntary. We choose 10 content images and 10 style images to synthesize 10 stylized images in each method and then ask subjects to select their favorite one. By the end of this user study, we had collected 341 votes from these anonymous participants. As shown in Figure 8, we show the percentage of votes for each method. The result shows that the stylization results obtained by our method are more appealing than those of other methods.

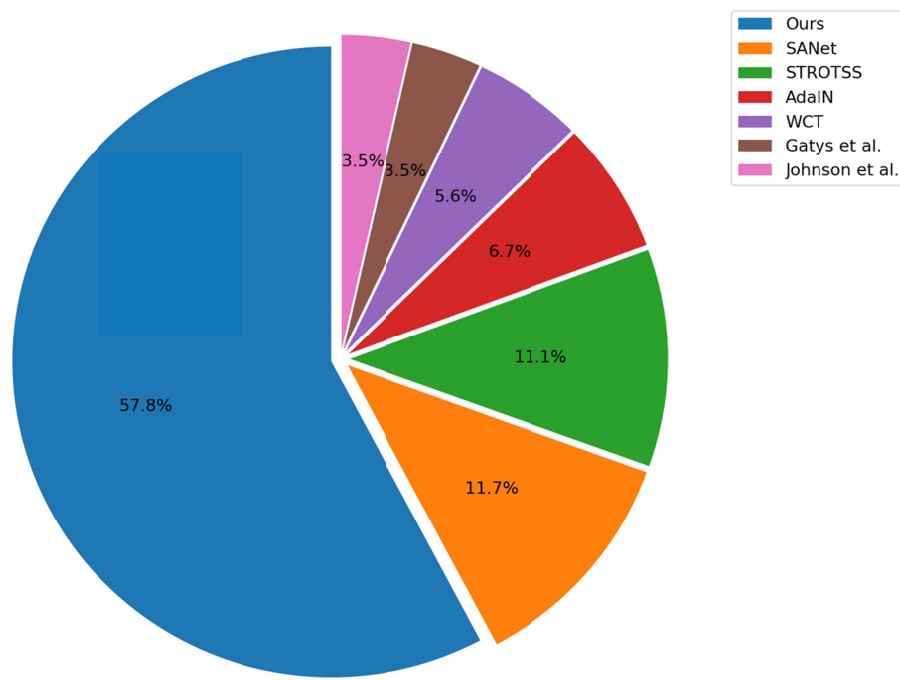

**Figure 8.** User preference results of our method and six state-of-the-art methods.

*4.6. Ablation Study on Loss Function*

We conduct ablation experiments to verify the effectiveness of each loss term used for training our model, and the results are shown in Figure 9. (1) Without perceptual loss $l_p$, too many structures of the content image are discarded; for example, the basic structure of the dog disappears in the stylized image. (2) Without gram matrix loss $l_g$, the stylization result is acceptable because mean-variance loss $l_m$ has a similar effect to $l_g$, but the color distribution of the stylized image is slightly different from that of the style image. Moreover, the textures of the dog in the stylized image are increasingly denser and smaller. (3) Without rEMD loss $l_r$, the texture distribution is chaotic, and some visual artifacts occur in the stylized image. (4) Without mean-variance loss $l_m$, the global color distribution of the stylized image is not exactly the same as that of the style image; for example, the dark color of the dog in the stylized image is more similar to that in the content image. This dark black color is completely absent in the style image.

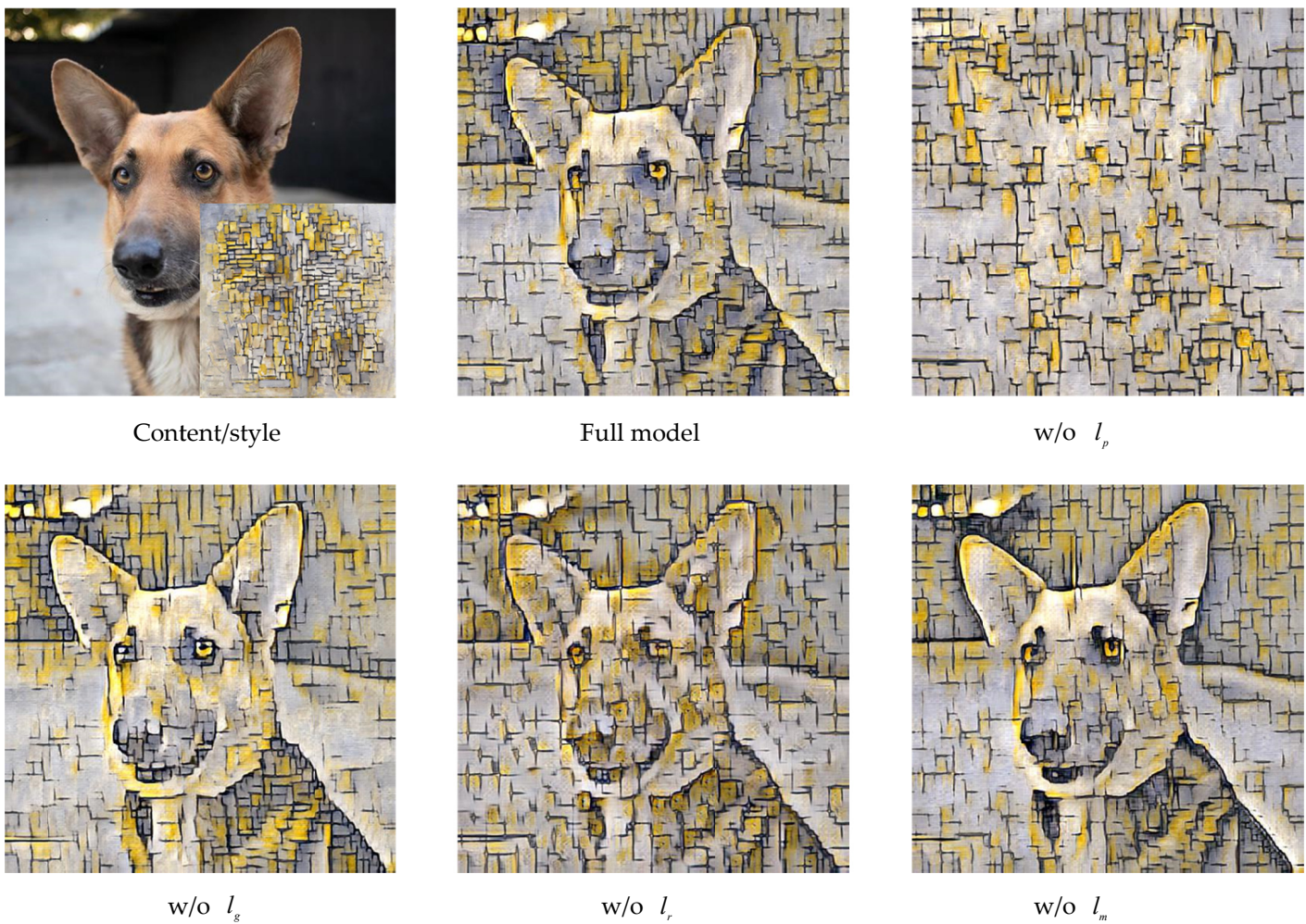

**Figure 9.** Ablation study of the effects of different loss functions used during training.

*4.7. Effectiveness of Coarse Network*

During training, we compare our full model with the model without the Coarse Network. As shown in Figure 10, our full model is trained faster than the model without the Coarse Network. The preliminary stylization result can be obtained with fewer iterations. Moreover, the stylized images of the comparison during the training phase are shown in Figure 11. At 3,000 iterations, our full model can generate a stylized image with a basic structure, while the model without the Coarse Network generates a completely unstructured image. At 10,000 iterations, the stylization result of our full model is substantially acceptable. However, the stylized result of the model without the Coarse Network is less than satisfactory as the main structure has not been generated. At 30,000 iterations, the model without the Coarse Network finally synthesizes the final stylized image, but some messy textures and unnatural structures appear in the stylized image. Compared to this compromised stylized result, our full model can generate an enhanced promising stylized result with more refined details, such as the brushstrokes of the cat's fur and eyes at 30,000 iterations, which are more delicate and finer than those at 10,000 iterations.

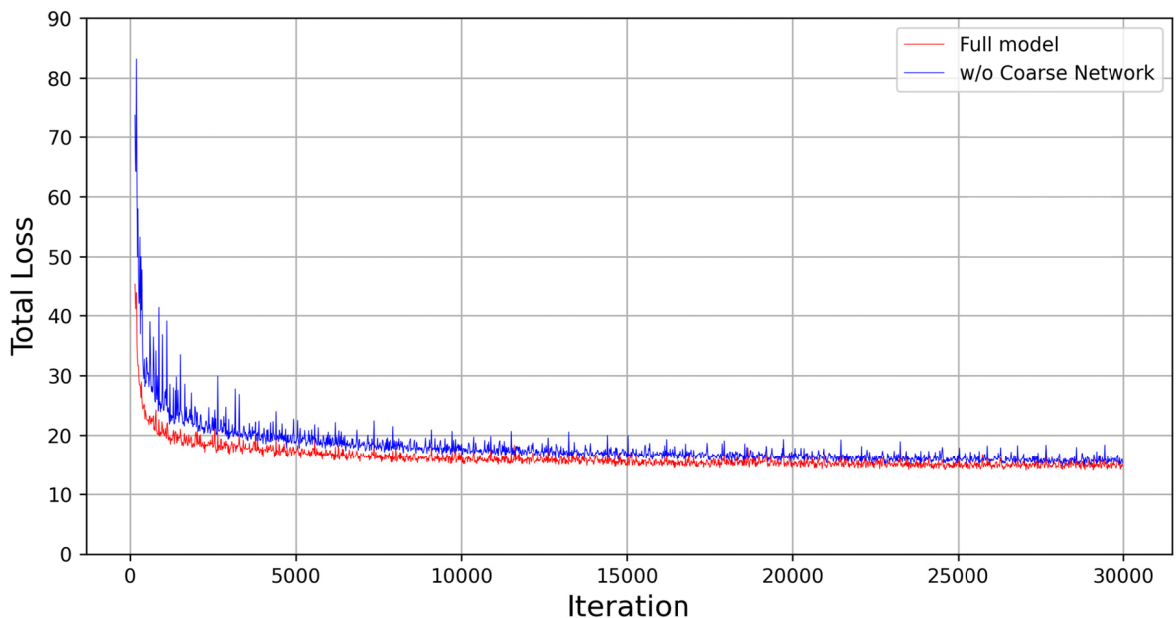

**Figure 10.** Comparison of the full model and the model without the Coarse Network in terms of total loss.

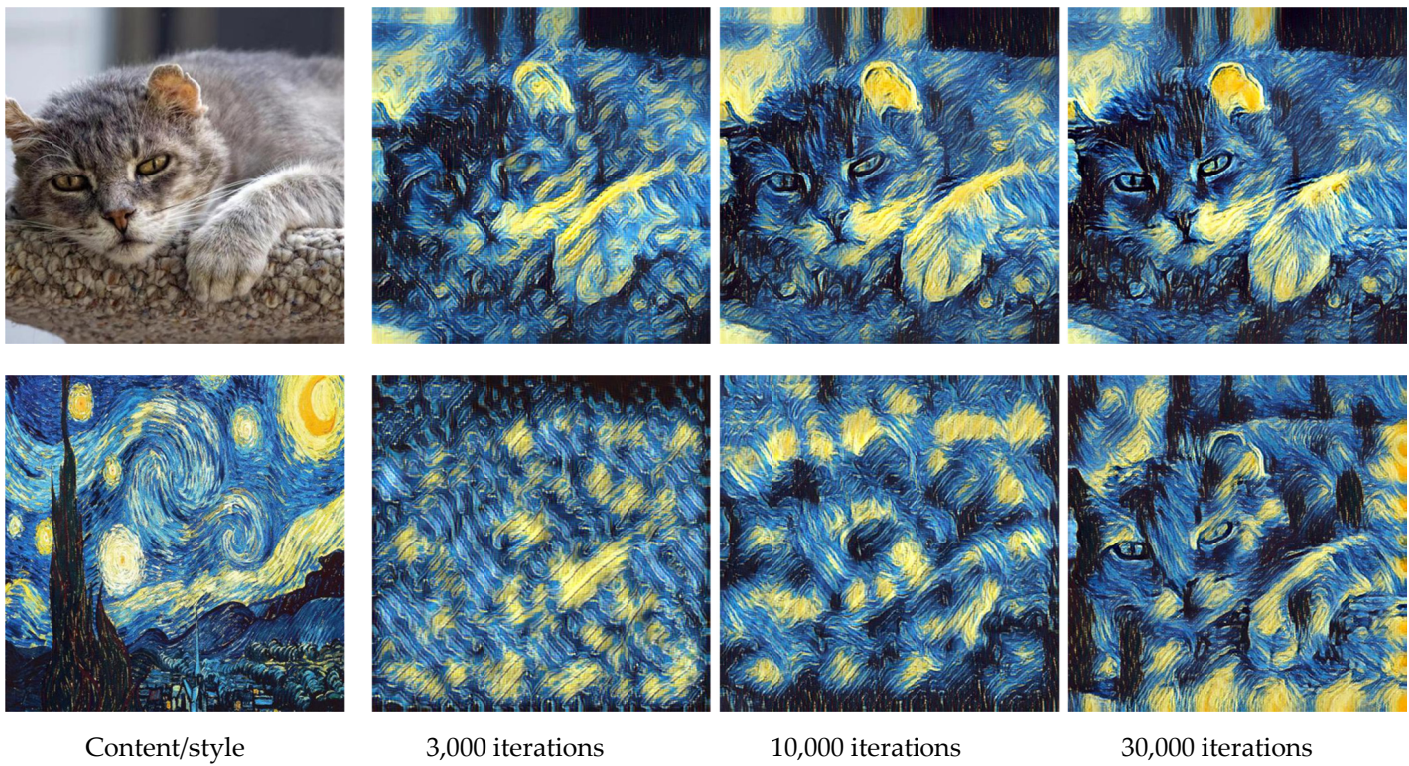

Content/style      3,000 iterations      10,000 iterations      30,000 iterations

**Figure 11.** Comparison of stylized images using the full model and the model without the Coarse Network during training. In the first row, the stylized results are generated by our full model. In the second row, the stylized results are generated by the model without the Coarse Network.

*4.8. Effectiveness of Fine Network*

As shown in Figure 12, we demonstrate the effectiveness of the Fine Network. Without the Fine Network, the Coarse Network can transfer the color and texture of style images, but the local details and global structure are worse than when our full model is utilized. The stylized image directly generated by the Coarse Network resembles an unfinished work in progress.

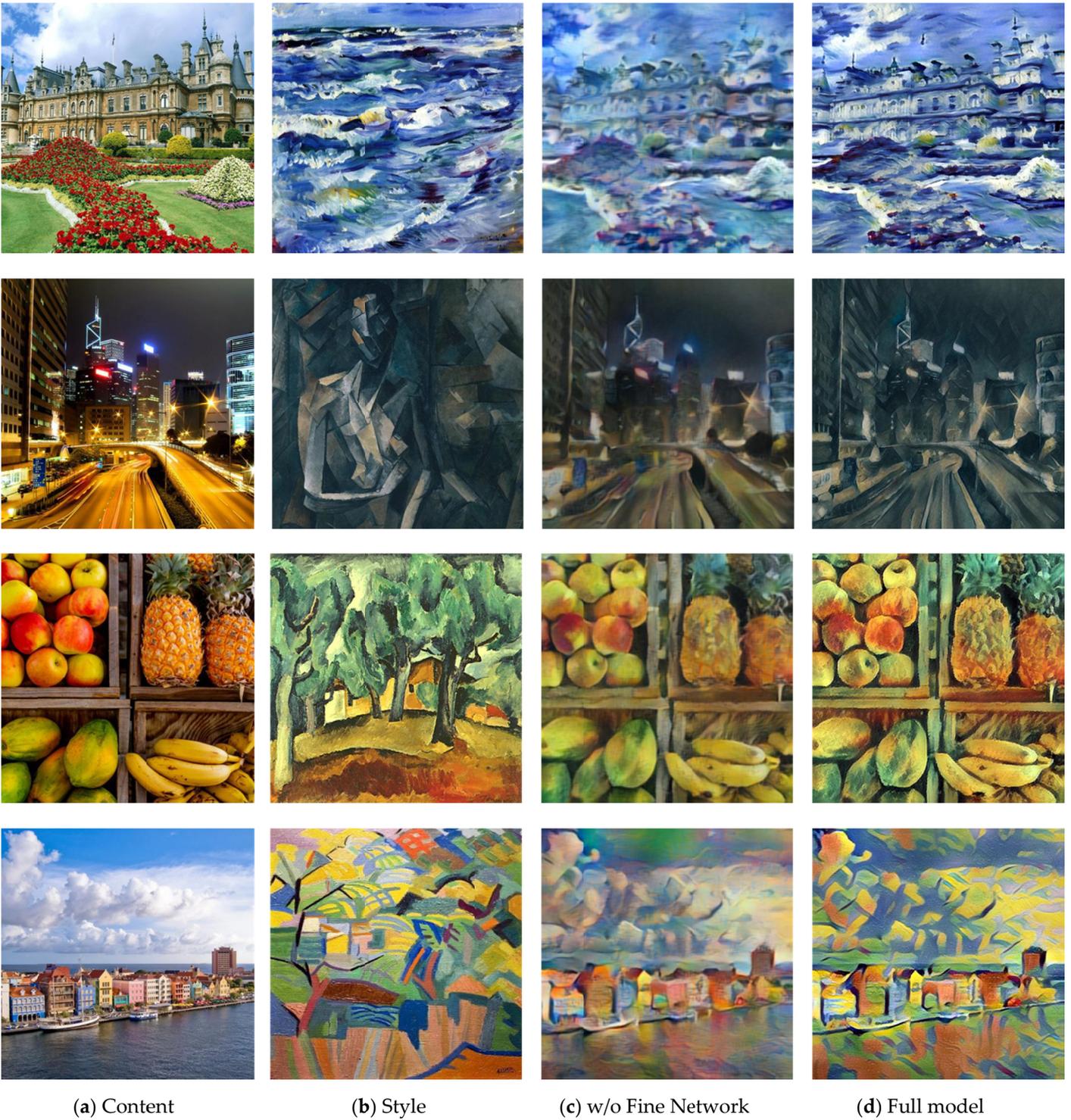

(**a**) Content  (**b**) Style  (**c**) w/o Fine Network  (**d**) Full model

**Figure 12.** Comparison of stylized images of the full model and the model without the Fine Network: (**a**) The content images; (**b**) The style images; (**c**) The stylized images generated by the model without Fine Network; (**d**) The stylized images generated by full model.

*4.9. Effectiveness of the SSF Modules.*

We compare two different feature fusion methods through some experiments. In the first method, the reconstructed coarse stylized features from the Coarse Network are fused into the reconstructed content features in the Fine Network based on our SSF modules. In the second method, we directly concatenate these two features for feature fusion. As Figure 13 shows, the stylization results based on the second method are transferred to the wrong color dis-tribution in some regions. Based on the first method, our model can accurately transfer the color distribution and more natural textures in the stylized images can be generated by selecting more critical information.

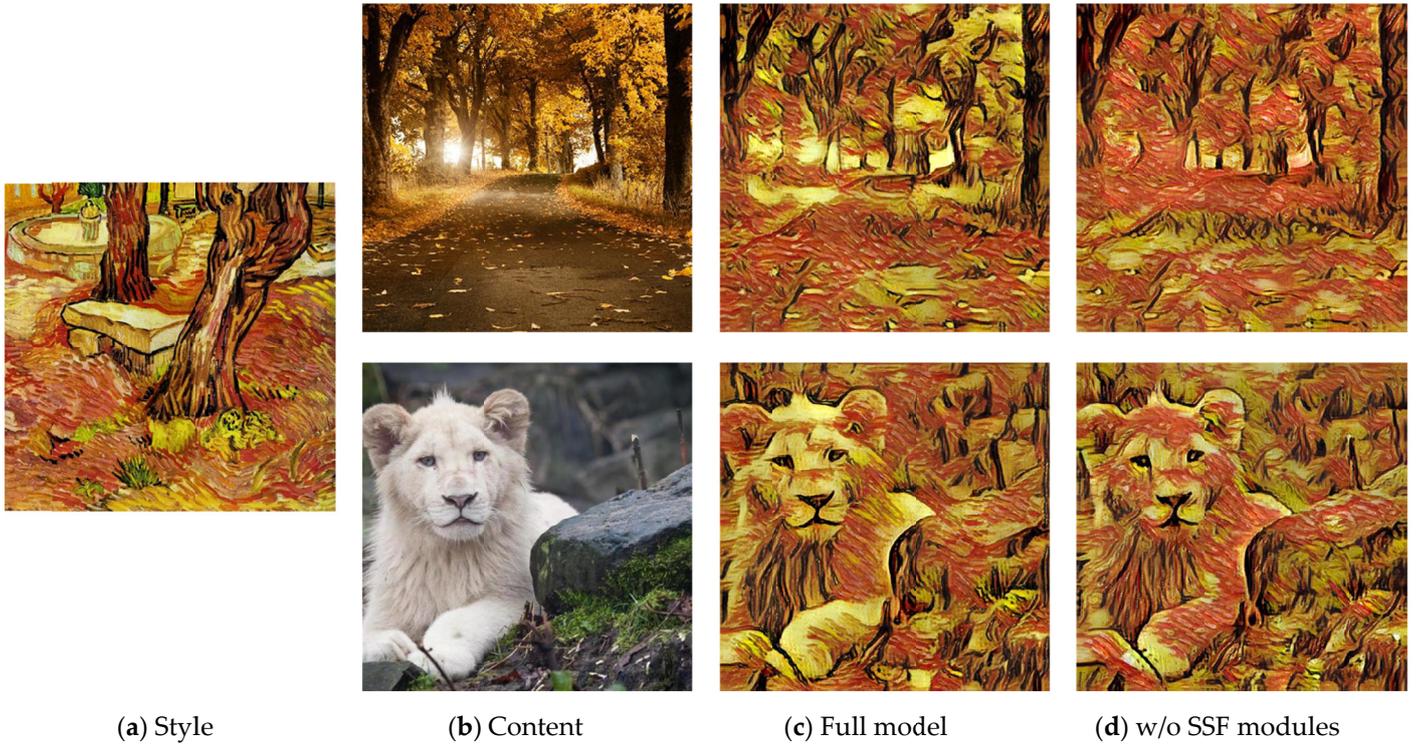

(**a**) Style  (**b**) Content  (**c**) Full model  (**d**) w/o SSF modules

**Figure 13.** Comparison of stylized images of our models with different feature fusion methods: (**a**) The content images; (**b**) The style images; (**c**) The stylized images generated by full model; (**d**) The stylized images generated by the model without SSF modules.

*4.10. Additional Experiments*

In Figure 14, we zoom in on some details in style images, content images, and stylized images. The local structures of these style images are transferred to the content image, and the object of the stylized images looks like a reasonable combination that is composed of the style structures rather than a simple mixture of the content structure and the style texture.

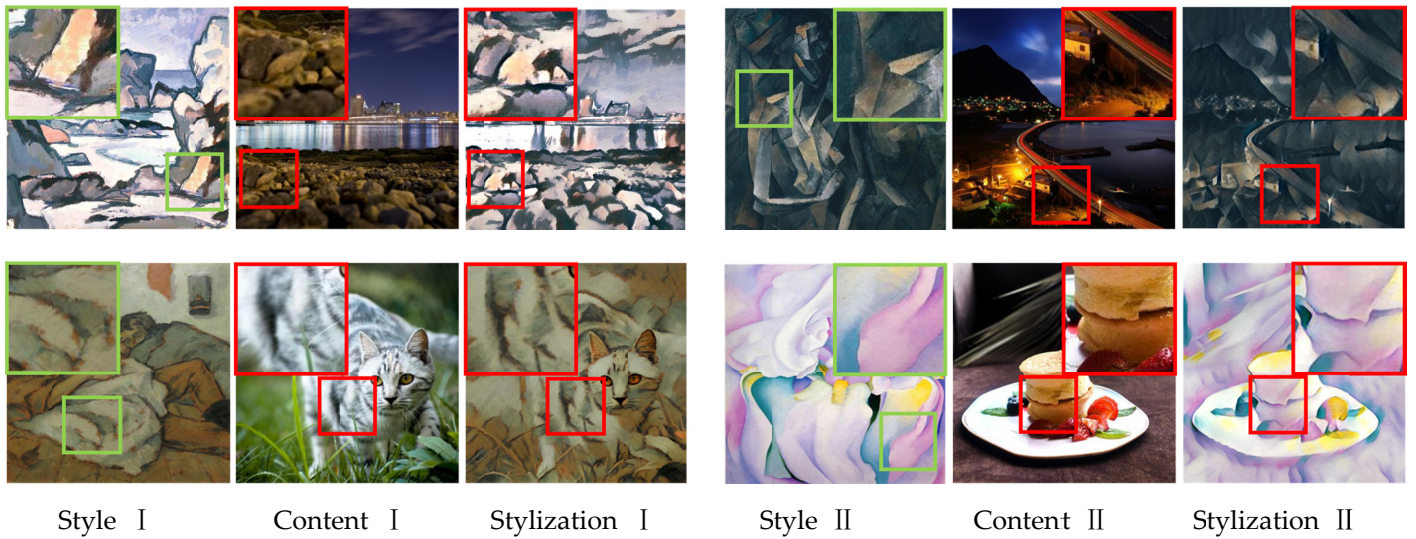

Style I  Content I  Stylization I  Style II  Content II  Stylization II

**Figure 14.** Comparison of local style details.

As shown in Figure 15, we can control the stylization degree by adjusting the weight term $\alpha$ in the training phase. It is demonstrated through these experiments that the main content structure can be preserved even though the stylization degree is large. Some local style structures, such as lines or color blocks, can be fused into the global content structure.

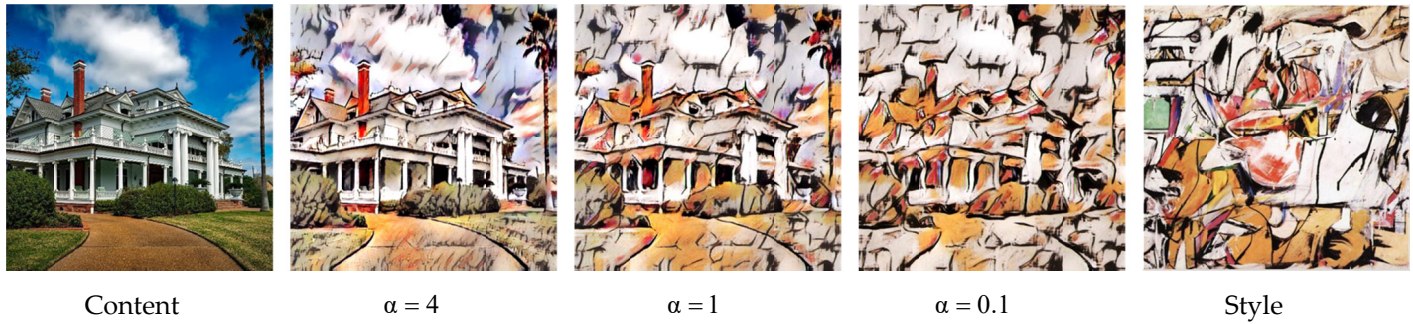

Content     α = 4     α = 1     α = 0.1     Style

**Figure 15.** Trade-off of content-style losses.

Following Gatys et al. [26], we incorporate color control and spatial control into our method. In Figure 16(**b**), the color distribution and the local structure of the stylized image are consistent with those of the style image. Then we use color control to make the stylized image preserve the global color of the content image. In Figure 16(**c**), although the color is similar to the content image, the local structure and texture are the same as those of the style image. In Figure 17, we use spatial control to transfer different regions of the content image to different styles. The stylization result is appealing as the local style structures and color distribution are maintained greatly. Both experiments demonstrate that our model can synthesize high-quality structure-aware stylized images by fusing key local structures from the style image into the main content structure while discarding some trivial details from the content image.

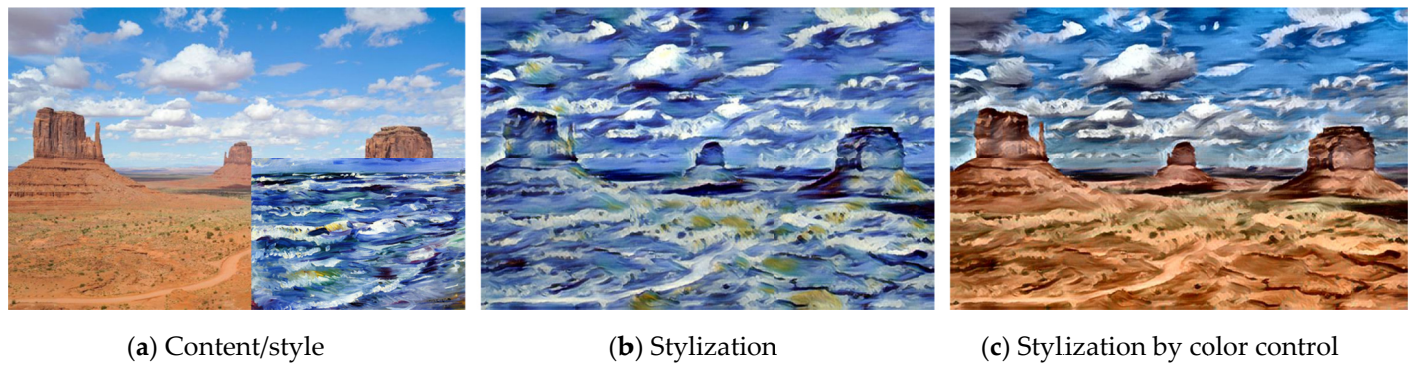

(**a**) Content/style     (**b**) Stylization     (**c**) Stylization by color control

**Figure 16.** Color control.

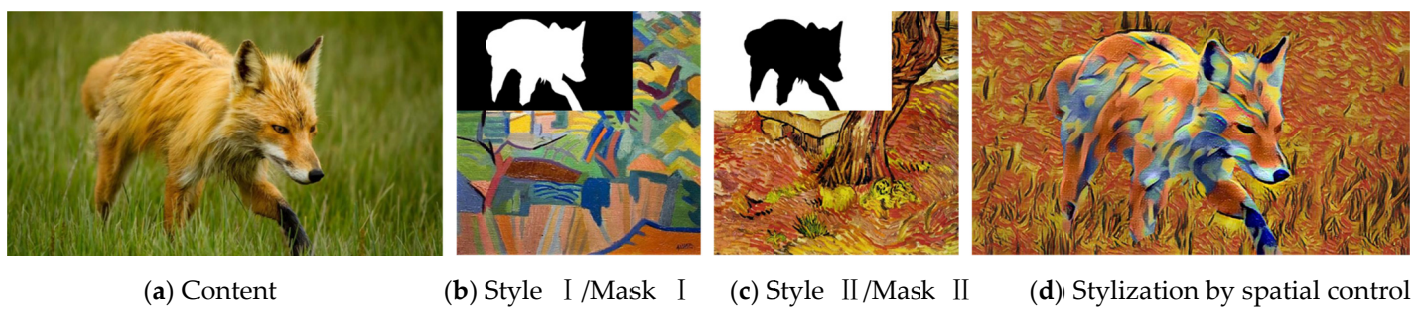

(**a**) Content     (**b**) Style Ⅰ/Mask Ⅰ     (**c**) Style Ⅱ/Mask Ⅱ     (**d**) Stylization by spatial control

**Figure 17.** Spatial control.

## 5 Conclusion

The conclusions are summarized as follows:

1.  We propose a novel feed-forward style transfer algorithm that fuses the local style structure into the global content structure. Different from most style transfer methods that work at the same scale, our model can integrate richer information of features from different scales and then synthesize high-quality structure-aware stylized images.
2.  We first propose a Coarse Network to generate reconstructed coarse stylized features at low resolution, which can capture the main structure of the content image and transfer the holistic color distribution of the style image. Then, we propose a Fine Network to enhance local style patterns and three SSF modules to selectively fuse the reconstructed stylized features into reconstructed content features at different levels.
3.  Through comparative experiments, it is demonstrated that our method is effective in synthesizing appealing high-quality stylized images, and these stylization results outperform the results generated by current state-of-the-art style transfer methods. The experimental results also demonstrate the effectiveness of the Coarse Network, the Fine Network, and the SSF module.

Although the high-quality stylization results can be synthesized by our method, our model only generates the stylized images with a single style after a training process. In future studies, we will achieve a novel arbitrary style transfer framework based on our full model in this paper. Appealing high-quality structure-aware stylized images with arbitrary style can be generated by this framework after a training process. In addition, we will try to use more feature transform methods to replace the whitening and coloring transforms for achieving higher running time efficiency.

**Author Contributions:** K.L. proposed the style transfer method and designed the framework. K.L. conducted the experiments. K.L. analyzed and discussed the experimental results. K.L. and G.Y. wrote the article. G.Y., H.W. and W.Q. revised the article and provided valuable advice for ablation experiments. All authors read and agreed to the published version of manuscript.

**Funding:** This research was funded by Natural Science Foundation of China (Grant No. 62162065, 62061049, 12263008), Application and Foundation Project of Yunnan Province (Grant No. 202001BB050032), Department of Science and Technology of Yunnan Province -Yunnan University Joint Special Project for Double-Class Construction (Grant No. 202201BF070001-005), Expert workstation of Yunnan Province（202105AF150011）, and Postgraduate Practice and Innovation Project of Yunnan University (Grant No. 2021Y177).

**Data Availability Statement:** The dataset of content images adopted during training is openly available online. MS-COCO: https://cocodataset.org/#home.

**Conflicts of Interest:** The authors declare no conflict of interest.

**Abbreviation List**

| | |
|---|---|
| SSF | Structural selective fusion |
| NPR | Non-photorealistic rendering |
| STROTSS | Style transfer by relaxed optimal transport and self-similarity |
| rEMD | Relaxed Earth mover's distance |
| AdaIN | Adaptive instance normalization |
| WCT | Whitening and coloring transforms |
| VGG | Visual geometry group |
| DualStyleGAN | Dual style generative adversarial network |
| POFMakeup | Peking Opera face makeup |
| SANet | Style-attentional network |
| LapStyle | Laplacian pyramid style network |
| ReLU | Rectified linear unit |
| GPU | Graphics processing unit |
| LPIPS | Learned perceptual image patch similarity |
| SSIM | Structural similarity index measurement |

**Symbol List**

| | |
|---|---|
| $x_c$ | Content image |
| $x_s$ | Style image |
| $x_{cs}$ | Stylized image |
| $\bar{x}_c$ | The result of downsampling $x_c$ by 2 |

| Symbol | Description |
|---|---|
| $\bar{x}_s$ | The result of downsampling $x_s$ by 2 |
| $\bar{f}_r^{(i)}$ | Restructured coarse stylized features |
| $c_r^{(i)}$ | Channels of $\bar{f}_r^{(i)}$ |
| $h_r^{(i)}$ | Height of $\bar{f}_r^{(i)}$ |
| $w_r^{(i)}$ | Width of $\bar{f}_r^{(i)}$ |
| $\bar{f}_c$ | Content feature extracted from VGG network |
| $\bar{f}_s$ | Style feature extracted from VGG network |
| $\bar{f}_c'$ | The result of linearly transforming $\bar{f}_c$ |
| $\bar{f}_{cs}$ | Stylized feature generated by WCT module |
| $f_{cs}$ | Reconstructed content features |
| $f_{csr}$ | The input of SSF module |
| $M_{cs}$ | Attention map of $f_{csr}$ |
| $f_{cs}'$ | The result of refining $f_{cs}$ |
| $f_{csr}'$ | The result of refining $f_{csr}$ |
| $f_{ssf}$ | The output of SSF module |
| $l_{re}$ | Reconstruction loss |
| $I_i$ | Input image |
| $I_o$ | Output image |
| $\Phi$ | VGG encoder that extracts features at ReLU_X_1 |
| $\lambda$ | Weight term of $l_{re}$ |
| $F_c^{(t)}$ | Content feature extracted at ReLU_$t$ |
| $F_s^{(t)}$ | Style feature extracted at ReLU_$t$ |
| $F_{cs}^{(t)}$ | Stylized feature extracted at ReLU_$t$ |
| $l_p$ | Perceptual loss |
| $l_r$ | Relaxed Earth mover's distance (rEMD) loss |
| $C$ | Cost matrix |
| $D_{cos}$ | Cosine distance |
| $l_g$ | Gram matrix loss |
| $G$ | Calculation of the Gram matrix |
| $l_m$ | Mean-variance loss |
| $\mu$ | Mean |
| $\sigma$ | Covariance |
| $L$ | Overall optimization objective |
| $\alpha$ | Weight term of $L$ |
| $\lambda_1$ | Weight term of $L$ |
| $\lambda_2$ | Weight term of $L$ |
| $\lambda_3$ | Weight term of $L$ |